\documentclass{article}

\usepackage{arxiv}

\usepackage[utf8]{inputenc} 
\usepackage[T1]{fontenc}    
\usepackage{hyperref}       
\usepackage{url}            
\usepackage{booktabs}       
\usepackage{amsfonts}       
\usepackage{nicefrac}       
\usepackage{microtype}      
\usepackage{lipsum}
\usepackage{graphicx}
\usepackage{algorithm}
\usepackage{algorithmic}
\usepackage{amsmath,amssymb,amsfonts}
\usepackage{graphicx}
\usepackage{textcomp}
\usepackage{xcolor}
\usepackage{multicol}

\title{Speedup deep learning models on GPU by taking advantage of efficient unstructured pruning and bit-width reduction}

\author{
 Marcin Pietro\'n \\
  Institute of Electronics\\
  AGH-UST\\
  Cracow, Poland \\
  \texttt{pietron@agh.edu.pl} \\
   \And
 Dominik \.{Z}urek \\
  Institute of Computer Science\\
  AGH-UST\\
  Cracow, Poland \\
  \texttt{dzurek@agh.edu.pl} \\
}

\begin{document}
\maketitle
\begin{abstract}
This work is focused on the pruning of some convolutional neural networks (CNNs) and improving theirs efficiency on graphic processing units (GPU) by using a direct sparse algorithm. The Nvidia deep neural network (cuDnn) library is the most effective implementations of deep learning (DL) algorithms for GPUs. GPUs are the most commonly used accelerators for deep learning computations. One of the most common techniques for improving the efficiency of CNN models is weight pruning and quantisation. There are two main types of pruning: structural and non-structural. The first enables much easier acceleration on many type of accelerators, but with this type it is difficult to achieve a sparsity level and accuracy as high as that obtained with the second type. Non-structural pruning with retraining can generate a weight tensors up to $\sim90\%$ or more of sparsity in some deep CNN models.In this article the pruning algorithm is presented which makes it possible to achieve high sparsity levels without accuracy drop. In the next stage the linear and non-linear quantisation is adapted for further time and footprint reduction. This paper is an extended of previously published \cite{GPGPU_ACCELERATING} concerning effective pruning techniques and present real models pruned with high sparsities and reduced precision which can achieve better performance than the CuDnn library.
\end{abstract}


\section{Introduction}
Deep convolutional neural networks (CNNs) achieve some of the best results in various artificial intelligence tasks including image processing \cite{cnn_image_classification_1}\cite{cnn_image_classification_2}, \cite{cnn_object_detection_1} or natural language processing  \cite{kim-2014-convolutional}\cite{cnn_text_quant_1}\cite{cnn_nlp_1}. Deep neural networks for running training and inference process on large datasets need specialised accelerators such as GPGPUs or other dedicated hardware accelerators. Over the years, scientists have been looking for methods to accelerate the calculations of the convolution operation. The direct convolution algorithm to perform convolutions requires $N^2$ multiplications and \textit{N(N-1)} additions where \textit{N} is the size of the input. For the same input the Fast Fourier Transform (FFT) method reduces operation complexity to \textit{O(N$log_{2}$(N))} \cite{fft_improvment}. The Winograd algorithm is suitable for small fixed-size kernels and requires 2.25 times fewer multiplications than direct convolution \cite{winograd_lavin}. The convolution operation can be realised by matrix-multiplication  \cite{conv_mm}, especially on the GPGPU which is highly tuned for performing this operation \cite{gmmem_gpu}. The NVIDIA deep neural network library (cuDNN)\footnote{https://developer.nvidia.com/cudnn} performs convolution with different algorithms (Winograd, FFT and GEMM) depending upon filter size, batch size and data representation. Apart from choosing different algorithms for speeding up convolution there are some other algorithms based on complexity and memory footprint reduction. Some CNN models for image processing, natural language processing or other task can be heavily pruned and theirs weights and activations bit-width can be reduced.
Depending upon the level of sparsity, it can be worth performing the convolution through the application of the direct sparse convolution method proposed by \textit{Chen} \cite{chen2018escoin} and extended by \textit{\.{Z}urek et al.} \cite{GPGPU_ACCELERATING}. The work concentrates on two main issues. The first is about methods for complexity reduction. It explains pruning and quantisation methodologies on chosen CNN models and theirs results. The second is focused on investigating when it is worth using sparse operations, instead of using dedicated NVIDIA libraries to run the convolution layer. As the main optimization strategy we propose the introduction of a unified sparse level for each of the output channels in each convolutional layer. The second strategy is determining the most optimal number of thread blocks for each convolutional layer separately. The presented approach is optimized towards the optimal arrangement of the data in order to obtain speedup with the direct convolution approach using the sparse format. These steps are crucial for achieving optimal efficiency. This work shows real examples of CNN models where it is possible to obtain the high level of sparsity so that acceleration using the presented algorithm vs cuDnn could be possible. These examples are well known CNN models like VGG and Resnet-50 architectures used on smaller and less complex data sets. The high sparsity levels were obtained by the presented pruning approach. Apart from achieving a high sparsity the accuracy levels were also improved. Finally, the impact on time efficiency of using half precision (FP-16) in a direct sparse convolution is explored. It is compared with cuDnn, where for 16-bit data representation, NVIDIA Tensor Cores specialised arithmetic units are used. In the presented work both linear and nonlinear approaches of quantisation were applied on  pretrained and pruned models. To our knowledge, this is the first work that shows the acceleration of the unstructured sparsity of weights compared to the dense approach using real models with highly optimized direct sparse convolution algorithm.

\section{Related work}
Convolution efficiency optimization have recently become quite popular research subject. Jord\'a \textit{et al.} \cite{cudnn_comaprision} present the way in which the cuDnn library calculates convolution layers dependent upon parameter configurations and data representation. Lavin \textit{et al.} \cite{winograd_lavin} introduces Winograd convolution implementation which is based on minimal filtering algorithm. This approach for a small filter and batch size was 2.26 times faster than the previous version of cuDnn. Ad\'amek \textit{et al.} \cite{fft_improvment} proposes an FFT based convolution on GPGPU by the shared memory implementation of the overlap-and-save method, and for certain sizes, a 30\% speed increase was achieved in comparison to cuDnn. The direct sparse convolutions method was proposed in \cite{Liu_dircect_sparece_idea}. The authors used the CSR format to store the weights and perform the convolution operation by use of the sparse matrix multiplication. This approach achieved 3.1-7.3 times speed increase comparison to dense convolution in the AlexNet model, on Intel Atom, Xeon and Xeon Phi processors. \textit{Lu et al.} \cite{sparse_winograd_fpga} proposed FPGA's sparse convolution implementation which in VGG16 is almost three times faster than FPGA's dense implementation. The same type of convolution was applied on the GPU in \cite{chen2018escoin}, where the speed increase for AlexNet \cite{AlexNet_architecture}, GoogleLeNet \cite{GoogleLeNet_architecture} and ResNet \cite{ResNet_Architecture} models were respectively 1.74, 1.34 and 1.43 times that of the GEMM implementation in the CUBLAS\footnote{https://developer.nvidia.com/cublas} library. \textit{\.{Z}urek et al.} \cite{GPGPU_ACCELERATING} introduced improvements to the direct sparse convolutions method which enabled the achievement of better performance than the cuDnn library in some specific cases of sparsity level. The comparison of results of the proposed method with the cuDnn library method was conducted for 32-bit and 16-bit representations of the data. The VGG-16, CNN-non-static and 1x1 layers from
ResNet models were used as benchmarks. The greatest accelerations were achieved for convolution type 1D, where for sparsity level equaling around 90 \% the speed was doubled. For the remaining types of convolution with the same sparsity level, depending on the layer type, the acceleration level was around twelve percent. An important role in sparse convolution is played by weight pruning, which can produce a number of zero weights. Information about the level of weight sparsity can be used after the pruning step in order to decide if it is worth running direct sparse implementation or cuDnn. 
In \cite{pietron2020retrain}, the authors prove that retraining with pruning can reduce the drop in accuracy caused by removing unimportant weights.
Pruning is one of the most popular solutions when it comes to memory compression and the acceleration of deep learning models \cite{pietron2020retrain}, \cite{frankle2019}, \cite{movement_pruning2020}. 
Some of the most popular approaches of pruning methods are: pruning without retraining with local search heuristics \cite{pietron2020retrain},\cite{motaz2020}, lottery tickets search \cite{frankle2019}, movement pruning  \cite{movement_pruning2020} and \cite{10.1145/3007787.3001163} or based on variational dropout \cite{molchanov}. The methods based on retraining use masks (static or dynamic) in each epoch which are mainly based on absolute values of weights or gradients. In most of the mentioned works there is no real use of the results obtained from unstructured sparsity in the GPU. In \cite{chen}, \cite{hu}, \cite{li}, \cite{liu}, \cite{wang}, \cite{zhuang} and \cite{aketi} research work concentrates on structural pruning. The approaches presented in these papers are focused mainly on pruning whole channels in a filters by using different techniques. In the following articles \cite{huang}, \cite{zhao}, \cite{molchanov}, \cite{frankle2019}, \cite{movement_pruning2020} algorithms for unstructured pruning are described without any hardware accelerator mappings.  
Quantisation is the next step by which it is possible to reduce workload and memory further. Many quantisation approaches were applied for deep learning    
\cite{han2015learning} \cite{linda2016} \cite{gysel2016ristretto} using linear or nonlinear quantisation, regularization modifications, clustering \cite{pietronCANDAR} and other techniques \cite{motaz2020}.

\section{Framework for deep learning compression}


In this section, the framework for deep learning model compression is described - UnSparse-Opt. The UnSparse-Opt compression framework consists of two main components: pruning and quantisation module.
First, the pruning algorithm based on retraining
produces layers with certain number of zero weights. Then these layers are mapped to direct sparse implementation. In the next phase, quantisation incorporated with linear and nonlinear bit width reduction is applied. Finally, efficiency of layers with reduced number of weights and bit format is compared with CuDnn. Based on these results, the network is configured to be partially run with the direct sparse approach. This process is described in Algorithm \ref{alg:main}.

\begin{algorithm}
\begin{algorithmic}[1]
\REQUIRE{sparsity thresholds for specific layers}
\STATE{pruning}
\STATE{quantisation}
\STATE{comparing pruned and quantized layers efficiency with cuDnn}
\STATE{network configuration}
\caption{Main scheme applying reduced bit format and unstructured sparsity in GPU}
\label{alg:main}
\end{algorithmic}
\end{algorithm}

\subsection{Pruning algorithm}

The proposed pruning method is based on retraining. Pruning with retraining guarantees much better final sparsity levels. The algorithm incorporates the evolutionary technique with rewinding during its execution. Rewinding allows to assign to not pruned weights values which they have before last training epoch \cite{renda2020}. The evolutionary part consists of genetic operators that can mutate and cross over the model with random solutions from an online built ranked list.
The input parameters of the Algorithm \ref{alg:pruning} are: 
$\epsilon$ - threshold for accuracy changes, $N_{it}$ - number of iterations of the algorithm, $\alpha$ - constant for weights update in mask increment process, $R_L$ - length of ranked list of best solutions, $[\Delta_0, \Delta_1,...,\Delta_N]$ - vector of units of sparsity by which the pruning is done in a layer in single step, $\gamma$ - threshold for mutation and crossover. 
The novelty of this pruning approach is that it combines few features of other algorithms all together (rewinding \cite{renda2020}, gradient analysis \cite{movement_pruning2020}) in one evolutionary approach. The algorithm prunes the model incrementally by using the dynamic layers sensitivity. Therefore, it differs from genetic approach in \cite{xu2021} where constant masks are used.


The representation of the pruned model $F_{\Theta}^{p}$ 
is the following tuple:
\begin{equation}
F_{\Theta}^{p} = (F_{\Theta}, M, O, G).
\label{eq:pruned_model}
\end{equation}
where $F_{\Theta}$ is the original model with set of convolutional and fully-connected layers $f_{\theta_{i}}$, which are executed with specified order (eq. \ref{eq:model}):
\begin{equation}
F_{\Theta}(X) = {f_{\theta_L}(f_{\theta_{L-1}}...(f_{\theta_{0}}(X)))},
\label{eq:model}
\end{equation}
$M$ is the masks set (eq. \ref{eq:mask_}):
\begin{equation}
M = \{M_{0}, M_{1},...,M_{L}\},
\label{eq:mask_}
\end{equation}
$O$ is the set of the optimizer parameters (type of the optimizer and learning rate), $G$ is the gradient statistics tensor. 


The weights of convolutional and fully-connected layers are represented by $\Theta$ (eq.\ref{eq:theta_}) tensor:

\begin{equation}
\Theta = \{\theta_{0}, \theta_{1},...,\theta_{L}\},
\label{eq:theta_}
\end{equation}

The mask is a binary tensor which consists of '0' and '1' elements. Each mask $M_{i}$ is assigned to specific convolutional or fully-connected layer and has the same shape $SH_{\theta_{i}}$=$C_{i}\times D_{i}\times H_{i} \times W_{i}$ as its weights tensor (eq. \ref{eq:mask__}, $C_i$ - number of input channels, $D_i$ - number of output channels, $H_i$ - filter height, $W_i$ - filter width, in case of fully connected layer: $W_i\gets1$ and $H_i\gets1$). 

\begin{equation}
M_{i} \in \{0,1\}^{SH_{\theta_{i}}}
\label{eq:mask__}
\end{equation}

The size $S_{\theta_i}$ of ${\theta_{i}}$, $M_{i}$ and $G_{i}$ is equal to $C_i\cdot D_i\cdot H_i \cdot W_i$. The number of zero elements in a mask is equal to the sparsity level (eq. \ref{eq:mask___}):

\begin{equation}
sparsity_i=\frac{S_{\theta_i}-\sum_{j=0}^{S_{\theta_i}} M_{i}[j]}{S_{\theta_i}}.
\label{eq:mask___}
\end{equation}

The weighted sparsity is computed as the sum of products in which each multiplication consists of size of the layer and its sparsity. The sum is divided by the number of all weights in the network (eq.\ref{eq:size} and \ref{eq:w_sparsity}):

\begin{equation}
S = \sum_{j=0}^{L}C_j \cdot D_j \cdot H_j \cdot W_j
\label{eq:size}
\end{equation}

\begin{equation}
sparsity_{ws} = \sum_{j=0}^{L}\frac{C_j \cdot D_j \cdot H_j \cdot W_j \cdot sparsity_j}{S}
\label{eq:w_sparsity}
\end{equation}


The Algorithm \ref{alg:pruning} starts from scratch with random initial weights (alg. \ref{alg:pruning}, line 1). The masks are initiated in random way with starting sparsity level after model weights initialization (Algorithm \ref{alg:pruning}, line 2).
Then, algorithm initiates sensitivity list (line 3), ranked list with $R_L$ copies of the initial model $F_{\Theta}$ (line 4), $\Delta$ and sparsity vectors (line 6-7).
In each iteration $t$, some subset of $l$ layers is chosen for further pruning (line 10). This step helps to gather statistics about layers sensitivity and diagnoses which layer may inhibit the learning process. The batch training is then performed (line 11). After each batch, the chosen layer weights are set to zero using the element wise multiplication with the layer mask (eq. \ref{eq:w_mask}):

\begin{equation}
\begin{split}
   \theta_{i} = \theta_{i} \odot M_{i}
   \end{split}
   \label{eq:w_mask}
\end{equation}


After the batch training, gradient analysis is performed (line 12). The gradient statistics are measured during the training. The whole training process is batch partitioned and each batch training bases on stochastic gradient-descent. The tensor $G$ stores mean values of weight gradients during the one training epoch. It expresses the importance of weights during learning process (eq. \ref{eq:gradient}, $B$ is number of batches in an epoch, $X_j$ is the $j$-th batch, $t$ is the iteration number):

\begin{equation}
G_{i}^t=\frac{1}{B}\sum_{j=0}^{B} \frac{\delta F_{\Theta}(X_j) }{\delta \theta_{i}}
\label{eq:gradient}
\end{equation} 

After that, the accuracy of current configuration is measured on a validation set (line 13), then sensitivity is computed (line 14). Based on this, sensitivity and delta parameter of the chosen layers $l$ are updated (line 16) and (line 17, eq. \ref{eq:delta}). If the sensitivity is negative the $\Delta_i$ is decreased. The updating factor in eq.\ref{eq:delta} is scaled by $\beta$ parameter. In presented experiments $\beta$ is 10. 
 
\begin{equation}
\Delta_i = \Delta_i + \frac{sensitivity_i \cdot \Delta_i}{\beta} 
\label{eq:delta}
\end{equation}

The sensitivity is compared with $\epsilon$ parameter (line 18). If the current model accuracy is better than than lowest accuracy in models ranked list, the pruned model is written to the ranked list set (line 19-21). In the ranked list $\Theta$, $M$, $G$ and accuracy of the model are written. Next, the worst solution from ranked list is removed (line 22). The ranked list has constant length, which is equal to $R_L$ parameter.
Then mask is updated (line 24). First, $g_i[c,d,h,w]$ importance indicators for all weights are computed. They are the weighted sum of absolute weight value and absolute average magnitude of its gradient (eq. \ref{eq:w_update}). The $\alpha$ parameter defines how important these factors are. In all experiments $\alpha$ parameter is set to 0.1. The gradient statistics help not only focus on weights magnitudes but also the dynamics of their changes. 



\begin{equation}
   g_i[c,d,h,w] = \alpha \cdot abs(G_{i}[c,d,h,w])+(1-\alpha) \cdot abs(\theta_i[c,d,h,w])
   \label{eq:w_update}
\end{equation}

In the next step the threshold parameter $\epsilon_S$ is set. The value of $\epsilon_S$ should guarantee new sparsity level which is computed by adding $\Delta_i$ to the layer sparsity value (eq. \ref{eq:threshold}): 
\begin{equation}
sparsity_i^{t} = sparsity_i^{t-1} + \Delta_i
\label{eq:threshold}
\end{equation}

Finally, the mask is updated by eq. \ref{eq:mask_def}:
\begin{equation}
M_{i}[c,d,h,w] =
  \begin{cases}
    0  &   \quad \text{if } g_{\theta_{i}}[c,d,h,w] < \epsilon_S\\
    1  & \quad \text{if } g_{\theta_{i}}[c,d,h,w] > \epsilon_{S}
  \end{cases}
\label{eq:mask_def}
\end{equation}





If the progress in training the model is not satisfactory (line 25) the mutation or crossover is performed (line 26-33). Finally, rewinding is executed (line 34). Mutation is just a random sparsity change by $\alpha_m \cdot \Delta_i$ in a previously chosen layers (line 28 and 29). The sparsity of the new mask in mutated layer is received by subtracting scaled $\Delta_i$ value from its previous sparsity level (eq.\ref{eq:mask_mutation}): 

\begin{equation}
sparsity_i^{t} = sparsity_i^{t-1} - \alpha_m \cdot \Delta_i
\label{eq:mask_mutation}
\end{equation} 

Based on the sparsity change, the $\epsilon_S$ parameter is recomputed and the layer mask is updated by eq. \ref{eq:mask_def}. The new mask $M_{l}'$ is received. 
Finally, the new mutated model is evolved by the element wise multiplication (eq. \ref{eq:w_mask}) of the weights with the new mutated masks in $l$ layers (eq.\ref{eq:mutation_func}):



\begin{equation}
M' = \{M_{0}, M_{1},...,M_{l}',...,M_{L}\}
\label{eq:mutation_func}
\end{equation}


The crossover takes the random solution from ranked list ($R[r]$). Next, it gets its mask ($M^{R[r]}$). Finally, the masks of layers $l$ are inserted into mask $M$ of the current solution and give new mask $M'$ (eq. \ref{eq:crossover_function} and \ref{eq:cross_func}, line 31 and 32).  



\begin{equation}
cross(M, M^{R[r]}, l) \rightarrow M'
\label{eq:crossover_function}
\end{equation}

\begin{equation}
M' = \{M_{0}, M_{1},...,M_{l}^{R[r]},...,M_{L}\}
\label{eq:cross_func}
\end{equation}













\begin{algorithm}
\begin{algorithmic}[1]
\REQUIRE{$\epsilon$, $\alpha$, $\Delta$, $\gamma$, $R_L$, $N_{it}$}
\STATE{$F_{\Theta}$ $\gets$ init model}
\STATE{$M$ $\gets$ init mask}
\STATE{$SL$ $\gets$ init sensitivity list}
\STATE{$R$ $\gets$ init ranked list}
\STATE{$L'$ $\gets$ $\{0,1,...,L\}$}
\STATE{$R'$ $\gets$ $\{0,1,...,R_L\}$}
\STATE{$\Delta$ $\gets$ set initial $\Delta_i$ values}
\STATE{$sparsity$ $\gets$ set initial $sparsity_i^0$}
\FOR{$t = 1$ \TO $N_{it}$}
\STATE{sample $l$ from L'}
\STATE{$F_{\Theta}$ $\gets$ train batches}
\STATE{$G$ $\gets$ compute gradient statistics}
\STATE{$accuracy^t$ $\gets$ compute accuracy of $F_{\Theta}$}
\STATE{$sensitivity^t$ $\gets$ $accuracy^t$-$accuracy^{t-1}$}
\STATE{$accuracy^{t-1}$ $\gets$ $accuracy^t$}
\STATE{$SL$ $\gets$ update layers sensitivity}
\STATE{$\Delta$ $\gets$ update delta}
\IF{$sensitivity^t$ $>$ $\epsilon$}
\STATE{$F_{\Theta}'$ $\gets$ get worst model from ranked list}
\IF{$accuracy^t$ $>$ accuracy of $F_{\Theta}'$}
\STATE{R $\gets$ write pruned model $F_{\Theta}^p$ to ranked list}
\STATE{R $\gets$ remove the worst pruned model $F_{\Theta}'$ from ranked list}
\ENDIF
\STATE{$M$ $\gets$ increment\_mask($M$, $\Delta$, $\alpha$, $l$)}
\ELSE
\STATE{sample p from $\mathcal{N}(0,\,1)$}
\IF{p $>$ $\gamma$}
\STATE{sample $\alpha_m$ from $\mathcal{N}(0,\,1)$}
\STATE{$F_{\Theta}^p$ $\gets$ mutation($F_{\Theta}^p$, $l$, $\alpha_m$)}
\ELSE
\STATE{sample $r$ from R'}
\STATE{$F_{\Theta}^p$ $\gets$ cross($F_{\Theta}^p$, R[r], $l$)}
\ENDIF
\STATE{$F_{\Theta}$ $\gets$ rewinding($F_{\Theta}$, $M$)}
\ENDIF
\ENDFOR
\end{algorithmic}
\caption{Pruning - main algorithm}
\label{alg:pruning}
\end{algorithm}

\section{Quantisation}

After the process of network distillation by the pruning process, quantization can be performed as the next step of reducing model complexity. Quantization is the procedure of constraining values from a continuous set or more dense domain to a relatively discrete set. It is possible to define a general mapping from a floating-point data $x\in {\mathcal S}$ to a fixed-point $q\in\mathcal Q$ using a function $f_{\mathcal Q}: \mathcal S\rightarrow \mathcal Q$ as follows (assuming signed representation):
\begin{equation}
	\label{eq:quant}
	q = f_{\mathcal Q}(x) = \mu + \sigma \cdot \text{round}(\sigma^{-1}\cdot (x-\mu)). 
\end{equation}
In our case, $\mu=0$ and $\sigma = 2^{-{\bf frac\_bits}}$ where: 
\begin{equation}
	{\bf int\_bits}=\text{ceil}(\log_{2}(\max_{x\in{\mathcal S}} |x|)) 
\end{equation}

and 
\begin{equation}{\bf frac\_bits}={\bf total\_bits}-{\bf int\_bits}-1.
\end{equation}
The scaling factor $\sigma$ is essentially just a shift up or down. A drawback is that a great deal of precision may be lost if the distribution of the data set ${\mathcal S}$ is skewed by a large mean.
Yet another approach can define the number of integer and fractional bits to represent regions of a distribution that will represent a large percentage of the range.  
In these cases, there will be saturation of a small percentage of the data, such as outliers, through the quantization procedure which may or may not significantly affect the accuracy. To determine the effects of saturation, one can experiment with different saturation levels. Therefore, histogram analysis is used in UnSparse-Opt to analyze outliers and set the best levels of saturation for activation of quantization. In presented experiments the threshold for activation saturation was set to 0.99. It means that values higher than 0.99$\cdot max(A)$ are saturated to 0.99$\cdot max(A)$ ($max(A)$ is maximum activation value).

To compare linear quantization results, a nonlinear technique based on clustering was implemented. 
This approach goes through all fully-connected and convolutional layers in a loop and clusters
the weights using KMeans algorithm (Algorithm \ref{alg:quant}, line 4). The weights are clustered to specified number of clusters $\omega$. Then, to each weight in a layer the identifier of the cluster is assigned (line 5-7), which is the closest one to its original value. 
The clusters centroids are quantize to $\psi$ bit-width format (line 8-10, in presented experiments $\psi$ is 8-bit). Finally, the codebook is created in which weights to cluster centroids mappings are stored. During forwarding pass each original weight value is mapped to $w_q$, which is reduced cluster centroid representation (line 12). This approach gives additional memory compression. The similar approach can be found in \cite{pietronCANDAR}.

\begin{algorithm}
\begin{algorithmic}[1]
\REQUIRE{$\omega$, $\psi$}
\STATE{$L$ $\gets$ get model fc and conv layers}
\STATE{$c_q$ $\gets$ empty list} //initial list for quantised centroids
\FOR{$l$ $\textbf{in}$ $L$}
\STATE{$C$, $\theta_{ca}$ $\gets$ KMeans($\theta_l$, $\omega$) //$C$ is a list of centroids, $\theta_{ca}$ is a list of weights to centroids assignment}
\FOR{$w$ $\textbf{in}$ $l$}
\STATE{$w$ $\gets$ $\theta_{ca}$[$w$]}
\ENDFOR
\FOR{$c$ $\textbf{in}$ $C$}
\STATE{$c_q$[c] $\gets$
quantise\_centroids($c$, $\psi$) //eq.20-22}
\ENDFOR
\FOR{$w$ $\textbf{in}$ $l$}
\STATE{$w_q$ $\gets$ $c_q$[$w$]}
\ENDFOR
\ENDFOR
\caption{Nonlinear quantisation}
\label{alg:quant}
\end{algorithmic}
\end{algorithm}

\section{Convolution operation using a sparse operation on GPGPU}
\label{implemenation_desc}
To perform convolution, the approach which was proposed in our previous work is used \cite{GPGPU_ACCELERATING}. The input data are stored in NCHW format (batch size, channel, height, width). The weights are stored in the CSR format, which requires the building of three arrays: \textit{rowptr} ($RP_i$), \textit{coldix} ($\Lambda_i$) and \textit{weights} ($\theta_i$). In order to avoid calculating indexes from the input array, which will be used to perform convolution, the $\Lambda_i$ array is modified to store these pre-computed indexes \cite{park2016faster}. As proposed in the previous work \cite{GPGPU_ACCELERATING}, the number of non-zero elements is the same for each output channel ($RP_i[j+1] - RP_i[j]$ returns the same value for each \textit{j}-th output channel). This number is equal to the highest number of non-zero elements which occurred in the output channels after the pruning process. That standardised sparse level was gained by marking some zero values as "non-zero". The CSR format treats them as normal value.  
As a result of this, it is not necessary to extract the sparsity level separately for each output channel. This measure ultimately decreases the calculation time by about  $26\%$- $28\%$ depending upon convolution type.

The details of the calculating convolution through the use of a sparse operation on GPU are presented in Algorithm \ref{alg:conv_by_sparse}. Each single thread block calculates one output channel. Therefore, the indexes ($sh_{idx}$) from which elements from $\theta_i$ and $\Lambda_i$ tensors are calculated through the usage of the following formula:
\begin{equation}
       sh_{idx} = RP_{i}[t_{out_z}\mod D_i] + t_{out_y}\cdot Y_w  + t_{out_x}
\end{equation}
where $t_{out_x}$, $t_{out_y}$ and $t_{out_z}$ are $x$, $y$ and $z$ global coordinates of thread \cite{cuda_programing_paper},  and $Y_w$ is output width. Based on these indexes tensors are copied from global to shared memory (Algorithm \ref{alg:conv_by_sparse}, lines 6-8).
\begin{algorithm}
\begin{algorithmic}[1]
\REQUIRE{$sb_S$, $D_i$, $t_{out_x}$,  $t_{out_y}$,  $t_{out_z}$ $s_h$, $X_w$, $X_{size}$, $Y_w$ $w_{pad}$, $s_w$}
\STATE{$N_B \gets (b_S \cdot out_C) \div sb_S$}
\STATE{$X^{G}$ $\gets$ $X^{C}$} //copy $X$ from CPU to GPU
\STATE{$\Lambda^{G}$ $\gets$ $\Lambda_i^{C}$} //copy $\Lambda$ from CPU to GPU
\STATE{$\theta^{G}$ $\gets$ $\theta_i^{C}$} //copy $\Theta$ from CPU to GPU
\STATE{$RP^{G}$ $\gets$ $RP_i^{C}$} //copy $RP$ from CPU to GPU
\STATE{$sh_{idx} = RP^G[t_{out_z}\mod D_i] + t_{out_y}\cdot Y_w  + t_{out_x}$}
\STATE{$\Lambda^{S}$ $\gets$ $\Lambda^{G}$ $[sh_{idx}]$} //copy $\Lambda$ from global to shared memory of GPU
\STATE{$\theta^{S}$ $\gets$ $\theta^{G}$ $[sh_{idx}]$} //copy $\Theta$ from global to shared memory of GPU
\FOR{$j=0$ $\TO$ $N_{nz}$}
\STATE{$w^{r}$ $\gets$ $\theta^{s}[j]$   //weight from shared array into registers}
\STATE{$\lambda^{r}$ $\gets$ $\Lambda^{s}[j]$ //$shift$ value from shared $coldix$ $array$ into registers}
\STATE{$out^r$ $\gets$ $0$}
\FOR{$t=0$ $\TO$ $sb_S$}
\STATE{$t_{in_{p}}$ $\gets$ $X_p + t_{out_x} \cdot s_h \cdot (X_w + w_{pad})+ t_{out_y} \cdot s_w$}
\STATE{$x^r$ $\gets$ $X[t_{in_{p}} + \lambda^{r} + t \cdot X_{size}]$}
\STATE{$out^r$ $\gets$ $out^r + w^r$ $\cdot$ $x^r$}
\ENDFOR
\STATE{$out^G$ $\gets$ $out^r$}
\ENDFOR
\STATE{$out^C$ $\gets$ $out^G$ //copy $out$ results to CPU}
\end{algorithmic}
\caption{Calculating convolution using a sparse operation on GPU. \textbf{Symbols}: $X$
-input data; $X_{size}$- size of single sample of input data;  $\Lambda_i$, $\theta_i$ and $RP_i$ (coldix arrays, weights array and rowptr - array from CSR format);  $N_{nz}$ - number of non zero elements in each output channel;}
\label{alg:conv_by_sparse}
\end{algorithm}

   \begin{figure}[ht]
 \includegraphics[width=\linewidth,
  clip=true]{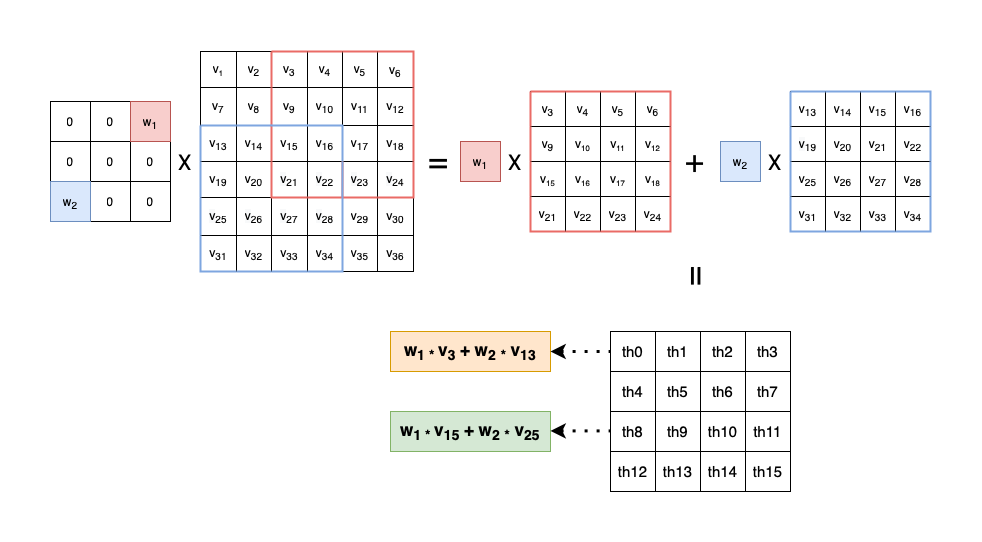}
  \caption{Calculating convolution using a sparse operation}
    \label{sparsce_conv_schema}
\end{figure}

At the beginning it is necessary to determine the sub-batch size ($sb_S$) value. This constant denotes the number of input vectors from the input batch which are handled by each single threads block (see Fig. \ref{numberOfBlock_schema}). Finally, the total number of running thread blocks is equal to: 

\begin{equation}
  N_B= \frac{b_S \cdot D}{sb_S}
  \label{eq:thread_blocks}
\end{equation}

 \begin{figure}[ht]
  \centering\includegraphics[width=\linewidth, viewport=0 400 566 775,
  clip=true]{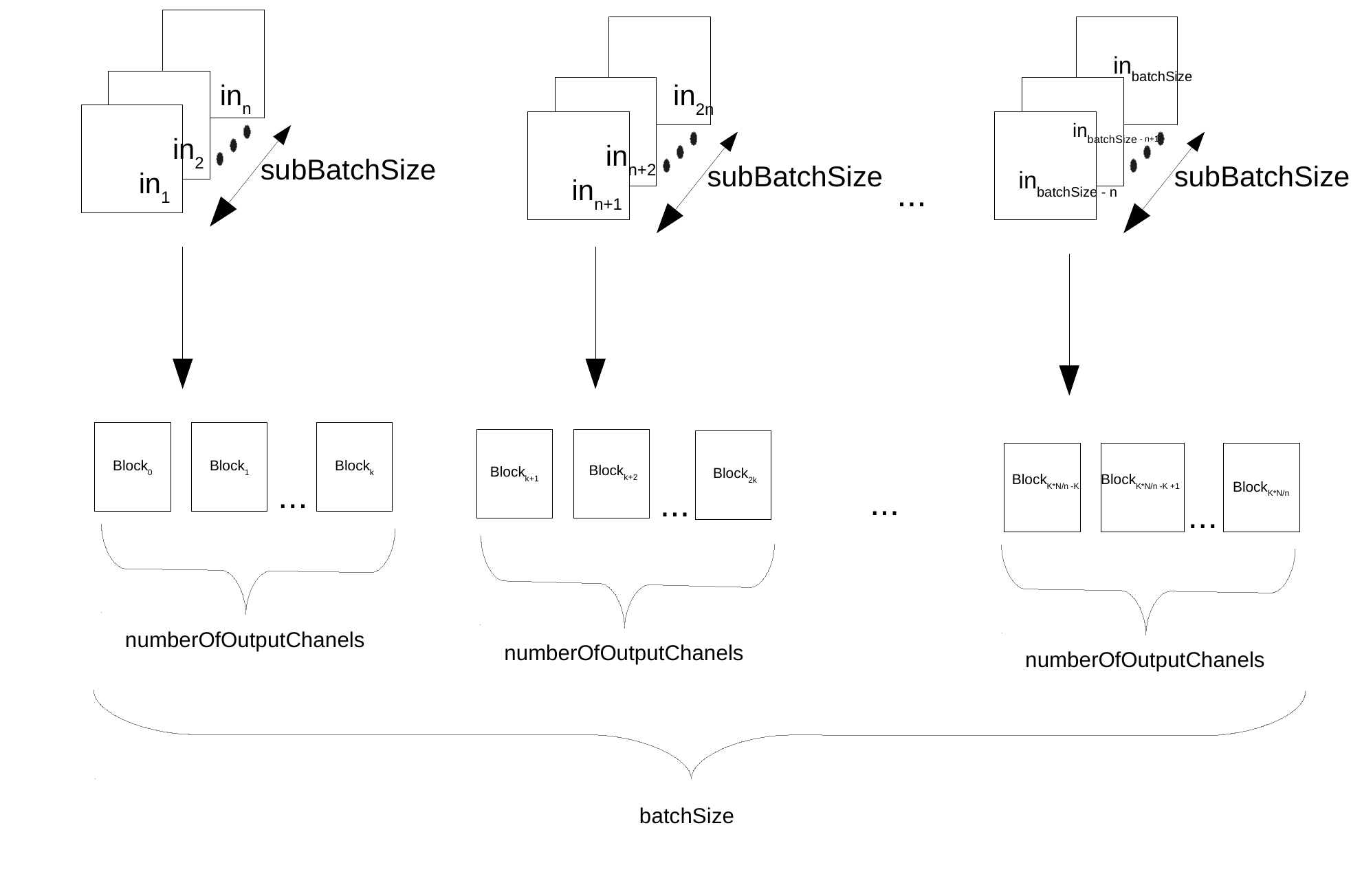}
    \caption{The total number of thread's block using to perform convolution for a single layer \cite{GPGPU_ACCELERATING}}
    \label{numberOfBlock_schema}
\end{figure}

 During the calculation of the convolution, non-zero values from the $\theta_i$ array and pre-calculated indices of the input vector from the $\Lambda_i$ array are loaded from shared memory into the thread local memory (Algorithm \ref{alg:conv_by_sparse}, lines 10 and 11). These values are reused for $sb_S$ input vectors (Algorithm \ref{alg:conv_by_sparse}, lines 13-17). This procedure enables maximum limitation of the reading from shared memory. Similar to the weights and indexes values, part of the input data is loaded from global memory and stored into registers (line 15). The addresses of input data ($t_{in_{p}}$), which need to be copied into registers, are calculated for each working thread though the usage of the following formula:
 
 \begin{equation}
 \begin{split}
    t_{{in}_{p}} = X_p + t_{out_x} \cdot s_h \cdot (X_w + w_{pad})+ t_{out_y} \cdot s_w  
     \end{split}
 \label{eq:load_in_to_registers}
 \end{equation}

 where $X_p$ is pointer to input array, $s_h, s_w$ are height and width of stride, $X_w$ is input width and $w_{pad}$ means width of padding.
 
 Each single thread is responsible for calculating one single output value by multiplication of the weight with a corresponding input value accumulating the partial sum and writing the final result to the global memory (Algorithm \ref{alg:conv_by_sparse}, lines 14-16 and 18). 
 This process is shown in Fig. \ref{sparsce_conv_schema}. In this case, the weights matrix contains only two non-zero values - $w_1$ (the red cell in the upper right corner of the 3x3 filter) and $w_2$ (the blue cell in the bottom left corner of the 3x3 filter). Both weights and input feature maps are marked as constants for CUDA kernel in order to hold them in the L2 cache, and coalesced memory access is provided \cite{cuda_programing_paper}.

 \begin{figure}[ht]
  \centering\includegraphics[width=\linewidth, viewport=0 400 566 775,
  clip=true]{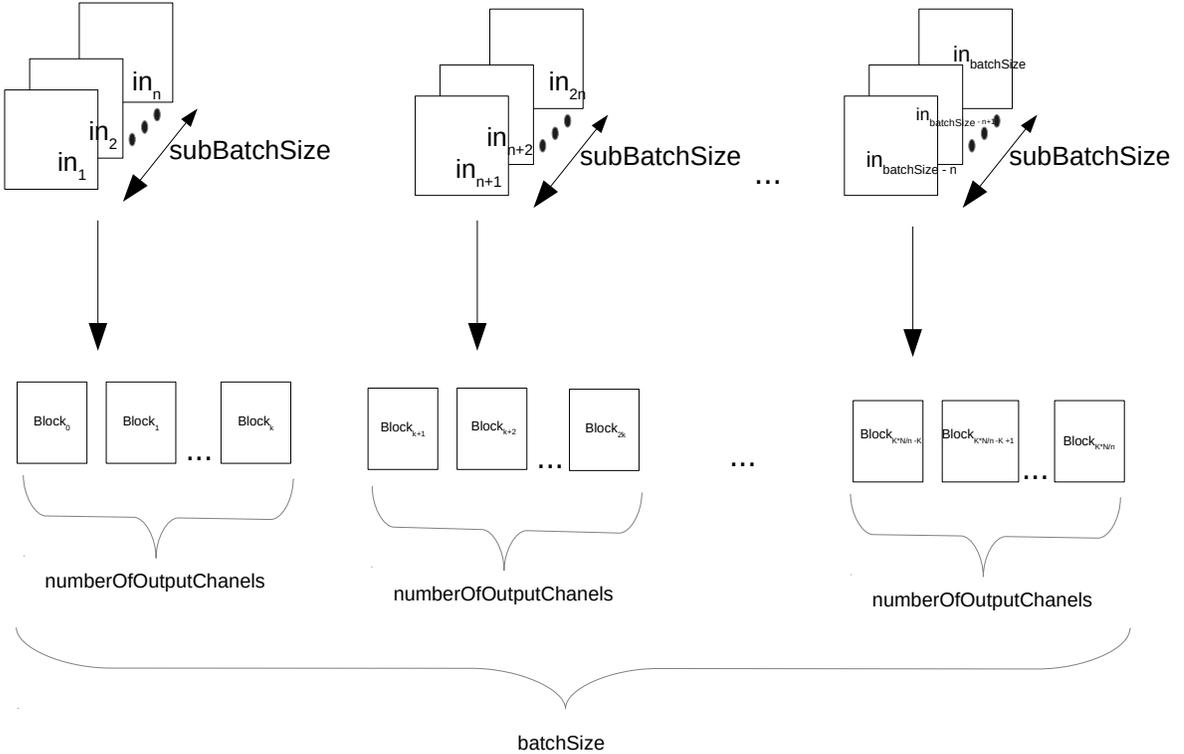}
    \caption{The total number of thread's block using to perform convolution for a single layer \cite{GPGPU_ACCELERATING}}
    \label{numberOfBlock_schema}
\end{figure}

 \begin{figure}[ht]
 \includegraphics[width=\linewidth,
  clip=true]{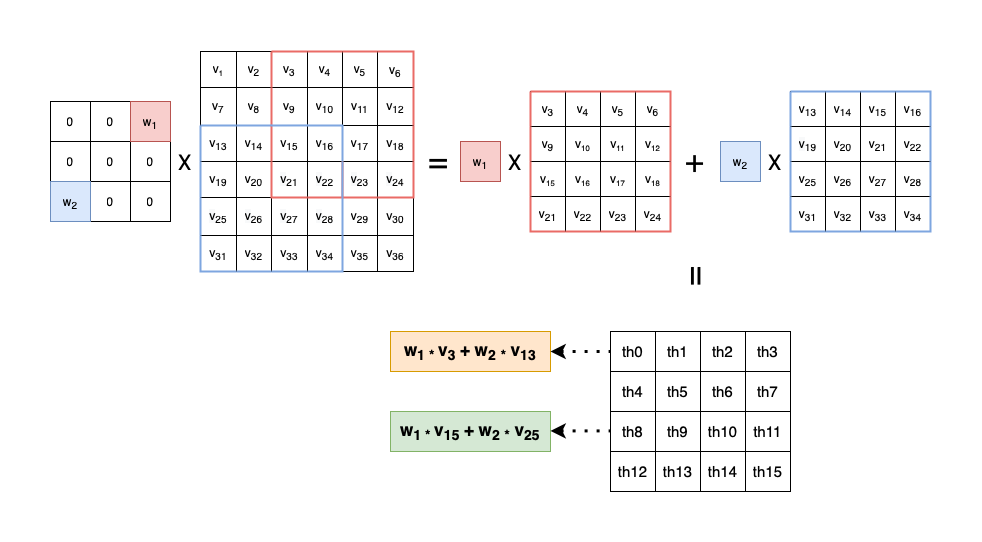}
  \caption{Calculating convolution using a sparse operation}
    \label{sparsce_conv_schema}
\end{figure}

\section{Results}
\label{result}
The proposed solution was verified in a number of experiments using the UnSparse-Opt framework. First, pruning was run on several models and datasets in different machine learning tasks. The accuracy of the pruned models was then tested. Next, quantisation effects were measured. Finally, the direct sparse convolution was performed on the models to compare this strategy with cuDnn efficiency.

\subsection{Pruning and quantisation}
\begin{table}[]
\centering

\begin{tabular}{|c|c|c|l|l|l|l|}
\hline
{\textbf{Method}} &{\textbf{\begin{tabular}[c]{@{}c@{}}Top1\\baseline $\rightarrow$ pruned \end{tabular}}} & \multicolumn{3}{c|}{{\textbf{\begin{tabular}[c]{@{}c@{}}Weighted Sparsity\\\textbf{ [\%]}\end{tabular}}}} \\
                                    &                                                                                & \multicolumn{3}{c|}{}                                                                                 \\ \hline
\textbf{UnSparse-Opt}                       & $90.01 \rightarrow 92.5 $                                                                         & \multicolumn{3}{c|}{96.2}                                                                            \\ \hline

\textbf{Li* \cite{li}}                       & $93.25 \rightarrow 93.4$                                                                          & \multicolumn{3}{c|}{64.0}                                                                         \\ \hline

\textbf{Liu* \cite{liu}}                      & $92.47 \rightarrow 91.75$                                                                          & \multicolumn{3}{c|}{36.0}                                                                            \\ \hline

\textbf{VCNN \cite{zhao}}                       & $93.25 \rightarrow 93.18$                                                                         & \multicolumn{3}{c|}{73.34}                                                                            \\ \hline

\textbf{Wang* \cite{wang}}                       & $93.13 \rightarrow 93.15$                                                                          & \multicolumn{3}{c|}{91.8}                                                                             \\ \hline

\textbf{StructuredBP \cite{neklyudov}}                       & $92.8 \rightarrow 92.5$                                                                          & \multicolumn{3}{c|}{75.17}                                                                             \\ \hline

\textbf{SparseVD \cite{molchanov}}                       & $92.45 \rightarrow 92.45$                                                                          & \multicolumn{3}{c|}{98.46}                                                                             \\ \hline

\textbf{Huang et al. \cite{huang}}                       & $92.77 \rightarrow 89.37$                                                                          & \multicolumn{3}{c|}{92.8}                                                                             \\ \hline

\textbf{DCP* \cite{zhuang}}                       & $93.99 \rightarrow 94.57$                                                                         & \multicolumn{3}{c|}{93.58}                                                                          \\ \hline

\end{tabular}
\caption{VGG16 pruning results on CIFAR10 (* - structural pruning method)}
\label{tab:vgg_pruned}
\end{table}
In this section, the results of the pruning and quantisation of the Resnet50, VGG16, FasterRCNN \cite{ren2015} and 1D CNN autoencoders \cite{faber2021} are shown. The models were run on CIFAR10, CIFAR100 and ImageNet (for classfication tasks), VOC Pascal (for object detection) and MSL (for anomaly detection).
In Tables \ref{tab:vgg_pruned} and \ref{tab:resnet_pruned_100}, the achieved level of sparsity by the pruning algorithm is described for VGG16 and Resnet50 on CIFAR10 and CIFAR100. The accuracy metric is top1. It measures the proportion of examples for which the predicted label matches the target label. The dense models are the baseline models (left side of $\rightarrow$ in Tables \ref{tab:vgg_pruned}, \ref{tab:resnet_pruned_100} and \ref{tab:resnet_pruned_imagenet}, the right side
of $\rightarrow$ is accuracy of pruned model). In the results, values for weighted sparsity are given. We can see that aside from high sparsity and huge memory reduction, the top1 accuracy is increased (error is reduced). In the case of Resnet50, it is noteworthy that the achieved pruned version is one of the smallest (memory footprint) of the TOP40 models run on the CIFAR100 (see ranking \cite{cifar100}). Additionally, a comparison of the results between the described pruning algorithm and others in the literature is presented. In the case of VGG16 run on CIFAR10 (Table \ref{tab:vgg_pruned}) the sparsity achieved by the algorithm is the second result after SparseVD \cite{molchanov}, but has a slightly better final accuracy (+0.05\%). Most of the methods in Table \ref{tab:vgg_pruned} are structural pruning-based algorithms. In the case of Resnet50 run on CIFAR100, our method outperforms other methods with regard to the level of sparsity (more than 90\%, Table \ref{tab:resnet_pruned_100}) and top1 accuracy. It is worth noting that accuracy after pruning of Resnet50 on CIFAR100 is one of the best achieved with this model without dataset augmentation (78.23\%). The VGG16 model run on CIFAR100 was pruned with 90\% sparsity level in all layers with very small drop in top1 (65.8\% (baseline) $\rightarrow$ 65.4\% (pruned)).

Our results were achieved by running 200 epochs of the training process with a ranked list length equal to 16. The batch size was set to 100 for CIFAR10 and to 128 for CIFAR100. The learning rate was 0.01. The VGG16 backbone in the FasterRCNN object-detection network was the next model compressed by UnSparse-Opt. The FasterRCNN feature extractor consists of thirteen VGG16 convolutional layers. The results are shown in Table \ref{tab:resnet_pruned_100}. The metric is mAP (mean average precision). Average precision computes the average precision value for recall value over 0 to 1 \cite{ren2015}. The mAP is the average of AP. The weighted sparsity is 85\% with a negligible drop in mAP (71.5\% (baseline) $\rightarrow$ 71.1\% (pruned)). 
In the third layer (of size 64x128), the eighth layer (of size 256x512) and all layers of size 512x512, it was possible to achieve a level of sparsity greater than or equal to 92\% without any significant drop in accuracy. This sparsity level guarantees direct sparse acceleration (see subsection 4.2).

\begin{table}[]
\centering

\begin{tabular}{|c|c|c|l|l|}
\hline
{\textbf{Method}} & {\textbf{\begin{tabular}[c]{@{}c@{}}Top1\\baseline $\rightarrow$ pruned \end{tabular}}} & \multicolumn{3}{c|}{{\textbf{\begin{tabular}[c]{@{}c@{}}Weighted Sparsity\\\textbf{ [\%]}\end{tabular}}}} \\
                                    &                                                                                & \multicolumn{3}{c|}{}                                                                                 \\ \hline

\textbf{UnSparse-Opt}                       & $67.06 \rightarrow 78.23$                                                                       & \multicolumn{3}{c|}{90.14}                                                                             \\ \hline

\textbf{Gradual Pruning* \cite{aketi}}                       & $71.48 \rightarrow 70.81$                                                                          & \multicolumn{3}{c|}{30}                                                                             \\ \hline

\textbf{SFP* \cite{hu}}                       & $74.29 \rightarrow 74.1$                                                                         & \multicolumn{3}{c|}{64}                                                                             \\ \hline

\textbf{Chen* \cite{chen}}                       & $70.01 \rightarrow 69.77$                                                                          & \multicolumn{3}{c|}{36.1}                                                                             \\ \hline
\end{tabular}
\caption{Resnet50 pruning results on CIFAR100 (* - structural pruning method)}
\label{tab:resnet_pruned_100}
\end{table}

The next models taken in the experiments were CNN autoencoders. The CNN autoencoders are one of the most efficient models used in anomaly detection tasks \cite{faber2021}. The simulations were performed on the Mars Science Laboratory (MSL) dataset \cite{nasa-lstm}. 
The dataset is separated into twenty-seven independent entities. In the presented experiments, one autoencoder architecture was trained and pruned separately on five chosen entities from the MSL dataset (M-2, M-6, F-7, D-16 and T-9). The model consists of twelve 1D convolutional layers (with 3x3 and 1x1 filters). Six of them are in the encoder and the next six are in the decoder module. The results of the pruned CNN autoencoders are shown in Table \ref{tab:auto_pruned}. In all cases, layers have been pruned to a sparsity level of 77\%.
The accuracy metric is $F1$. It is well-suited for anomaly detection tasks, as it is more resistant to class imbalance than other metrics such as top1 accuracy (which are susceptible to yield high values when the normal class in testing data is large and the anomaly class is underrepresented). 
As it is shown in Table \ref{tab:auto_pruned} the $F1$ score is still above or at the same level as it is in a dense baseline models. 

\begin{table}[]
\centering

\begin{tabular}{|c|c|c|l|l|}
\hline
{\textbf{Method}} & {\textbf{\begin{tabular}[c]{@{}c@{}}F1\end{tabular}}} & \multicolumn{3}{c|}{{\textbf{\begin{tabular}[c]{@{}c@{}}Weighted Sparsity\\\textbf{ [\%]}\end{tabular}}}} \\
                                    &                                                                                & \multicolumn{3}{c|}{}                                                                                 \\ \hline
                     & M-2                                                                     & \multicolumn{3}{c|} {}                                                                          \\ \hline
\textbf{baseline autoencoder-dense}                       &  0.78                                                                       & \multicolumn{3}{c|}{0.0}                                                                             \\ \hline

\textbf{pruned autoencoder}                       & 0.78                                                                       & \multicolumn{3}{c|}{77.0}                                                                             \\ \hline
 & M-6                                                                      & \multicolumn{3}{c|} {}                                                                          \\ \hline
\textbf{baseline autoencoder-dense}     &  0.29                                                                        & \multicolumn{3}{c|}{0.0}                                                                             \\ \hline

\textbf{pruned autoencoder}       &   0.29                                                                         & \multicolumn{3}{c|}{77.0}                                                                             \\ \hline
 & F-7                                                                      & \multicolumn{3}{c|} {}                                                                          \\ \hline
\textbf{baseline autoencoder-dense}     &  0.35                                                                       & \multicolumn{3}{c|}{0.0}                                                                             \\ \hline

\textbf{pruned autoencoder}       &   0.34                                                                         & \multicolumn{3}{c|}{77.0}                                                                             \\ \hline

 & D-16                                                                      & \multicolumn{3}{c|} {}                                                                          \\ \hline
\textbf{baseline autoencoder-dense}     &  0.33                                                                        & \multicolumn{3}{c|}{0.0}                                                                             \\ \hline

\textbf{pruned autoencoder}       &   0.34                                                                         & \multicolumn{3}{c|}{77.0}                                                                             \\ \hline
 & T-9                                                                      & \multicolumn{3}{c|} {}                                                                          \\ \hline
\textbf{baseline autoencoder-dense}     &  0.47                                                                        & \multicolumn{3}{c|}{0.0}                                                                             \\ \hline

\textbf{pruned autoencoder}       &   0.61                                                                        & \multicolumn{3}{c|}{77.0}                                                                             \\ \hline

\end{tabular}
\caption{Autoencoder pruning results on MSL dataset  \cite{nasa-lstm}}
\label{tab:auto_pruned}
\end{table}

\begin{table}[]
\centering

\begin{tabular}{|c|c|c|l|l|}
\hline
{\textbf{Method}} & {\textbf{\begin{tabular}[c]{@{}c@{}}Top1\\baseline $\rightarrow$ pruned\end{tabular}}} & \multicolumn{3}{c|}{{\textbf{\begin{tabular}[c]{@{}c@{}}Weighted Sparsity\\\textbf{ [\%]}\end{tabular}}}} \\
                                    &                                                                                & \multicolumn{3}{c|}{}                                                                                 \\ \hline

                     & Resnet50     &  \multicolumn{3}{c|}{}                                                                           \\ \hline
\textbf{UnSparse-Opt}                       &  76.13 $\rightarrow$ 74.51                                                                        & \multicolumn{3}{c|}{74.8}  \\
\hline                     

\textbf{UnSparse-Opt - 1x1 layers}                       &  76.13 $\rightarrow$ 68.0                                                                       & \multicolumn{3}{c|}{90.0 in 1x1 layers}  \\
\hline
\textbf{GenExp} \cite{xu2021}                       & 76.15 $\rightarrow$ 76.17                                                                          & \multicolumn{3}{c|}{~80.0}                                                                             \\ \hline
\textbf{DC} \cite{Han2016}                      &  76.15 $\rightarrow$ 76.35                                                                          & \multicolumn{3}{c|}{~70.0}                                                                             \\ \hline
\textbf{BEDL} \cite{han2018}                      &  76.15 $\rightarrow$ 76.35                                                                          & \multicolumn{3}{c|}{~80.0}                                                                             \\ \hline
\textbf{CFAR}  \cite{renda2020}                     &  76.15 $\rightarrow$ 76.35                                                                          & \multicolumn{3}{c|}{~80.0}                                                                             \\ \hline
                     & VGG16                                                                      & \multicolumn{3}{c|} {}                                                                          \\ \hline
                     \textbf{UnSparse-Opt}                       & 73.36 $\rightarrow$ 73.0                                                                       & \multicolumn{3}{c|}{95.0}  \\
\hline
\textbf{GenExp}  \cite{xu2021}                     &  69.2 $\rightarrow$ 69.25                                                                           & \multicolumn{3}{c|}{~95.2} \\ \hline
\textbf{ADMM-NN}   \cite{ren2019}                    &  69.0 $\rightarrow$ 69.0                                                                          & \multicolumn{3}{c|}{95.0} \\ \hline
\textbf{DC} \cite{Han2016}                &  71.33 $\rightarrow$  71.17                                                                        & \multicolumn{3}{c|}{~92.5} 
\\ \hline

\end{tabular}
\caption{Pruning results on ImageNet for Resnet50 and VGG16}
\label{tab:resnet_pruned_imagenet}
\end{table}
The results on ImageNet are described in Table \ref{tab:resnet_pruned_imagenet}. The results show that it is not possible to achieve levels of sparsities for Resnet50 and VGG16 which can give acceleration using a direct sparse approach (see subsection 4.2). The presented pruning algorithm and other well-known methods achieve weighted sparsities between 70\%-80\% (Table \ref{tab:resnet_pruned_imagenet}). In the case of sparsity levels which guarantee speedup in direct sparse (92\% in 3x3 filters and 90\% in 1x1 filters, see subsection 4.2), the drop of accuracy in Resnet50 is around 8-9\% (Table \ref{tab:resnet_pruned_imagenet}). In the case of VGG16, the weighted sparsity is quite high in all presented algorithms (92.5\%-95.2\%, Table \ref{tab:resnet_pruned_imagenet}). In VGG16, the largest layers are three fully-connected and they have the highest pruning scores. Some convolutional layers have sparsities slightly lower than 90\%, which makes them more time efficient in cuDnn than in the direct sparse method. The batch size used in experiments for ImageNet was 256. The learning rate was set to 0.01.

In Table \ref{tab:vgg_quant}  the results of quantisation approaches are presented. The linear sixteen-bit (half precision both for weights and activations - 16b/16b) quantisation and nonlinear cluster based four-bit quantisation were performed: weights are mapped to 16 centroids and each centroid represented in sixteen-bit format). The sixteen-bit quantisation called \textit{half} precision can be directly used in cuDnn as an option. In the case of four bit quantisation, it can be used in future hardware optimisations. The quantisation was run on pruned models. It is worth noting that in the case of the 16b/16b configuration, all models have no drop in accuracy (in many cases with an improvement above a 0.1\%, Table \ref{tab:vgg_quant}). In the case of 4b/16b, the 5.5\% average drop was observed on five tested 1D CNN auto-encoders. The 4b/16b does not heavily affect accuracy of CNN models on smaller datasets (VOC Pascal and CIFAR100), but it is more sensitive in the case of huge datasets like ImageNet (drop in accuracy about 6-7\%).
\begin{table}[]
\centering

\begin{tabular}{|c|c|c|c|c|}
\hline
{weights/activations} & {\textbf{CIFAR100}} & {\textbf{ImageNet}} & \textbf{VOC Pascal} & \textbf{MSL}\\
                                         
                                         \hline
\textbf{VGG16 *}                      & -1.7\%      & -7.1\%  & n/a          &    n/a                                                                                                                                  \\ \hline
\textbf{VGG16 **}                       & -0.1\%        &  +0.1\% & n/a     & n/a                                                                                                                                       \\ \hline

\textbf{Resnet50 *}                       & -3.6\%    & -6.1\% &   n/a     & n/a                                                                                                                                         \\ \hline

\textbf{Resnet50 **}                       & +0.1\%     & +0.1\% &  n/a    &    n/a                                                                                                                                       \\ \hline

\textbf{CNN-AE * (avg)}                       & n/a  & n/a & n/a             &    -5.5\%                                                                                                                                    \\ \hline

\textbf{CNN-AE ** (avg)}                       & n/a   & n/a &        n/a     &    +0.2\%                                                                                                                                  \\ \hline

\textbf{FasterRCNN *}                       & n/a  & n/a &      -1.9\%      & n/a                                                                                                                                         \\ \hline

\textbf{FasterRCNN **}                       & n/a   & n/a &    +0.1\%       &  n/a                                                                                                                                    \\ \hline

\end{tabular}
\caption{Accuracy drops in quantisation for Resnet50, VGG16, CNN-AE (autoencoder) and FasterRCNN (*4b/16b, **16b/16b).}
\label{tab:vgg_quant}
\end{table}

\subsection{Unstructured sparsity with reduced precision on GPU}
\label{subsec:gpu_result}


In Table \ref{tab:vgg_real}, the time efficiency of pruned and quantised VGG16 layers on CIFAR100 are compared with the cuDnn library. In the case of VGG16, despite such high sparsity levels, an improvement over the cuDnn was not achieved for every layer in the floating-point (Table \ref{tab:vgg_real}). The improvement was achieved for the last three VGG16 layers ($\sim14.4\%$ for \textit{float} and $\sim14.2\%$ for \textit{half} data type).
The direct sparse algorithm is always faster for the \textit{half} version than for \textit{float}. In the case of the cuDnn library, the \textit{half} convolution is always performed using the \textit{GEMM} algorithm which very often gives worse time efficiency than the \textit{float} version. In the layers with following input sizes: $64\times226\times226$, $256\times58\times58$ and $512\times30\times30$, the
\textit{half} approach is less efficient than \textit{float}.
In two types of VGG16 layers, cuDnn outperforms the direct sparse approach used with \textit{half} precision. These are layers of size 128x256 and 256x256 (Table \ref{tab:vgg_real}). In the rest of the layers, direct sparse with \textit{half} precision is most efficient option. 
In the case of the FasterRCNN model in the third layer (of size 64x128), the eighth layer (of size 256x512) and all layers with a number of input and output channels equal to 512x512, it was possible to achieve sparsity above 92\% without a drop in accuracy. It guarantees similar or better efficiencies in the floating precision of direct sparse in comparison to cuDnn. In \textit{half}, the performance of these layers in the direct sparse mode is significantly better. In the case of ImageNet, it is possible to achieve sparsities around 90\% in all $512x512$ layers (in the rest of the layers, sparsity is lower than 90\%). These results can guarantee some small improvements in \textit{half} precision in the last VGG16 layers. 
Having 90\% level of sparsity, it is possible to achieve better performance than the cuDnn for the 1x1 convolution. For this type of convolution, the cuDnn always uses the \textit{GEMM} algorithm. 
In the case of the ResNet50 run on CIFAR100 (Table \ref{tab:resnet_pruned_100}), it was possible to achieve sparsity levels of between 90\%-92.5\% in all 1x1 filters, which makes direct sparse faster than cuDnn. The results for 64x1x1x256 and 256x1x1x64 layers are included in Table \ref{tab:res_net_64}. The same levels of sparsities on ImageNet give a significant drop in accuracy (8-9\%). Therefore, it is not possible to achieve acceleration with direct sparse without significant drop in accuracy.

One dimensional convolutional layer needs the lower sparsity to be sped up using a direct sparse approach (Table \ref{tab:cnn_non_static_2}). In this case, the input data are in the shape of a vector, therefore to perform convolution by direct sparse method, less memory jumps are needed than with the 2D convolution. The 77\% of zero values in a kernel can give an improvement in time efficiency for layer of size 300x64 ($\sim9\%$ and $\sim11\%$ speed increase for the \textit{float} and \textit{half} respectively (Table \ref{tab:cnn_non_static_2}). The autoencoders incorporated with 1D convolutional layers used for multivariate anomaly detection can be pruned to such levels. In this case, significant speed up can be achieved (10\%-30\% time reduction to cuDnn). In all cases, the complexity in MAC operations is reduced by the sparsity ratio. The 16b/16b quantisation gives additional $4/9$ reduction. 

\begin{table}
\centering

\begin{tabular}{|c|c|c|c|c|c|} 
\hline
\begin{tabular}[c]{@{}c@{}}\textbf{Convolution size}\\\textbf{ (CHWD)}\end{tabular} & \begin{tabular}[c]{@{}c@{}}\textbf{Sparsity}\\\textbf{ [\%]}\end{tabular} & \begin{tabular}[c]{@{}c@{}}\textbf{Sparse}\\\textbf{ - float}\end{tabular} & \begin{tabular}[c]{@{}c@{}}\textbf{Sparse}\\\textbf{ - half}\end{tabular} & \begin{tabular}[c]{@{}c@{}}\textbf{cuDnn -~}\\\textbf{float~}\end{tabular} & \begin{tabular}[c]{@{}c@{}}\textbf{\textbf{cuDnn -~}}\\\textbf{\textbf{half}}\end{tabular}  \\ 
\hline
64x224x224x64                                                                       & 90                                                                        & 60.73                                                                           & 27.08                                                                          & 19.07                                                                           & 31.82                                                                                            \\ 
\hline
64x112x112x128                                                                      & 92                                                                        & 15.97                                                                           & 8.64                                                                           & 10.56                                                                           & 9.48                                                                                             \\ 
\hline
128x112x112x128                                                                     & 93                                                                        & 27,12                                                                           & 16.22                                                                          & 17.28                                                                           & 15.88                                                                                            \\ 
\hline
128x56x56x256                                                                       & 92                                                                        & 12.28                                                                           & 8.42                                                                           & 9.21                                                                            & 7.81                                                                                             \\ 
\hline
256x56x56x256                                                                       & 90.8                                                                      & 21.81                                                                           & 14.23                                                                          & 14.27                                                                           & 16.09                                                                                            \\ 
\hline
256x28x28x512                                                                       & 92                                                                        & 9.11                                                                            & 5.92                                                                           & 6.72                                                                            & 7.86                                                                                             \\ 
\hline
512x28x28x512                                                                       & 92                                                                        & 15.28                                                                           & 13.67                                                                          & 15.02                                                                           & 16.82                                                                                            \\ 
\hline
512x14x14x512                                                                       & 92                                                                        & 4.23                                                                            & 4.08                                                                           & 4.84                                                                            & 4.66                                                                                             \\
\hline
\end{tabular}

\caption{CuDnn and direct sparse time results [ms] for VGG16 }
\label{tab:vgg_real}
\end{table}

\begin{table}
\centering

\begin{tabular}{|c|c|c|c|}
\hline
\textbf{Data type} & \textbf{Layer size(CHWD)} & \textbf{cuDnn} & \textbf{Sparse convolution}  \\ \hline
\textbf{float}          &   64x1x1x256            & 0.35                                                                           & 0.32                                                                            \\ \hline
\textbf{half}           &  64x1x1x256            & 0.30                                                                         & 0.27                                                                        \\ \hline

\textbf{float}              &    256x1x1x64        & 0.37                                                                          & 0.31                                                                         \\ \hline
\textbf{half}       &    256x1x1x64                & 0.29                                                                        & 0.24                                                                         \\ \hline
\end{tabular}
\caption{Time results [ms] for ResNet50. 256 filters 1x1x64 and 64 filters 1x1x256 (sparsity $\sim90\%$), \cite{GPGPU_ACCELERATING}}
\label{tab:res_net_64}
\end{table}

The presented results were measured with the optimal value of running thread blocks (see section \ref{implemenation_desc}). This number strongly depends on the $sb_S$ parameter (see. Eq. \ref{eq:thread_blocks}), and the value of this parameter was empirically optimised for each layer and belongs to the set \{2, 4, 8\}. It could be observed that the optimal value is not universal for each layer. This means that this parameter depends on the size of the convolution. According to this, the optimal values of this parameter for older versions of GPU can be different because of the memory limitations (especially by the L2 cache size). In the case of the other parameters the efficiency ratio between direct sparse and cuDnn should be similar for different GPU architectures. 
Without setting this value by the method proposed in this paper, it would not be possible to achieve better performance than cuDnn. In the case of the number of thread blocks being equal to $b_S \cdot D$, for VGG16, ResNet50 the performance decreased by around $\sim10\%$. In the case of 1D CNN, the decrease was $\sim12\%$. An even larger decrease of performance occurred when all the data from the batch was processed by the $D$ blocks. This value was between $\sim38\%$ and $\sim45\%$. The exact influence of this parameter on particular layers is presented in Table \ref{tab:numberOfBlockInflueance}.

Both solutions, direct sparse convolution and the cuDnn library, were run through the usage of \textit{multiple CUDA streams} \cite{nvidia-link} but no time efficiency improvement was observed. This fact means that it is not possible to process two (or more) convolution operations concurrently on the same GPU with versions of CUDA and cuDnn used in described experiments.

Each of the presented experiments were performed on the Nvidia Tesla V100-SXM2-32GB \cite{v100-link} with the CUDA version 11.4  and cuDnn version 8.22. The batch size in the direct sparse method vs cuDnn comparison is always equal to 128. The final execution times are average values achieved from ten simulations. The standard deviation measured in run experiments was smaller than 1\%. 

\begin{table}
\centering

\label{tab:cnn_non_static_2}
\begin{tabular}{|c|c|c|c|c|} 
\hline
\begin{tabular}[c]{@{}c@{}}\\\textbf{Data type}\end{tabular} & \begin{tabular}[c]{@{}c@{}}\textbf{cuDnn time }\\\textbf{ \textbackslash{} algorithm}\end{tabular} & \multicolumn{3}{c|}{\begin{tabular}[c]{@{}c@{}}\textbf{Sparse convolution time}\\\textbf{~for given sparsity}\end{tabular}}  \\ 
\cline{3-5}
                                                             &                                                                                                    & \textbf{77\%} & \textbf{83\%} & \textbf{87,5\%}                                                                              \\ 
\hline
\textbf{float*}                                               & 0.192\textbackslash{}GEMM                                                                          & 0.176         & 0.126         & 0.102                                                                                        \\ 
\hline
\textbf{half*}                                                & 0.161\textbackslash{}GEMM                                                                          & 0.145         & 0.097         & 0.069                                                                                        \\
\hline
\textbf{float**}                                               & \multicolumn{1}{c|}{0.231\textbackslash{}GEMM}                                                    & 0.236         & 0.188         & 0.135                                                                                        \\ 
\hline
\textbf{half**}                                                & \multicolumn{1}{c|}{0.204\textbackslash{}GEMM}                                                    & 0.182         & 0.148         & 0.103                                                                                        \\
\hline
\end{tabular}
\caption{\label{tab:cnn_non_static_2}Time results [ms] for CNN 1D for input 300x64 (*kernel size 2x1, **kernel size 3x1) in CHWM format, \cite{GPGPU_ACCELERATING}}
\end{table}

\begin{table}
\centering
\begin{tabular}{|c|c|c|} 
\hline
\begin{tabular}[c]{@{}c@{}}\\\textbf{Layer type}\end{tabular} & \begin{tabular}[c]{@{}c@{}}\textbf{Number of running }\\\textbf{threads blocks}\end{tabular} & \begin{tabular}[c]{@{}c@{}}\textbf{Performance }\\\textbf{decrease}\end{tabular}  \\ 
\hline
\textbf{Resnet50 1x1}                                                  & $D \cdot b_S$                                                              & 10\%                                                                              \\ 
\hline
\textbf{Resnet50 1x1}                                                  & $D$                                                                        & 38\%                                                                              \\ 
\hline
\textbf{VGG16}                                               & $D \cdot b_S$                                                              & 10\%                                                                              \\ 
\hline
\textbf{VGG16}                                               & $D$                                                                        & 45\%                                                                              \\ 
\hline
\textbf{CNN 1D}                                       & $D \cdot b_S$                                                              & 12\%                                                                              \\ 
\hline
\textbf{CNN 1D}                                       & $D$                                                                        & 41\%                                                                              \\
\hline
\end{tabular}

\caption{Impact of the usage optimal number of the thread blocks on the performance}
\label{tab:numberOfBlockInflueance}
\end{table}

\section{Conclusions and future work}

This work is focused on pruning convolutional layers in various deep learning models and speeding them up on the GPU through the use of the sparse convolution algorithm. Additionally, the pruned models are quantised to reduce the complexity further. The UnSparse-Opt framework incorporated with pruning and quantisation algorithms is described to perform this process.
The work presents concrete cases when convolution using the direct sparse convolution can be more efficient than cuDnn. Most improvements are achieved for 1D convolution. 
The 1D CNN autoencoders are one of the described examples in which unstructured pruning with the direct sparse convolution can give a significant time reduction vs cuDnn implementation. It has been shown that 2D convolution using direct sparse convolution in some cases can also outperform cuDnn. This can be done on less complex datasets, where it is possible to achieve very high sparsity levels. The achieved results on CIFAR10 and CIFAR100 datasets show this phenomenon. 
In the case of large datasets like ImageNet, it has been shown that pruning algorithms can not achieve sparsity levels which can guarantee acceleration by using direct sparse method. 
Additionally, we examined the influence of conducting the calculation using reduced precision on time efficiency and model accuracy. 
It was shown that \textit{half} precision does not affect accuracy of the presented models. The four-bit weight quantisation is more sensitive on more complex datasets. The presented results show that direct sparse convolution with \textit{half} precision can in many cases outperform both \textit{half} and \textit{float} cuDnn versions.

Future work will further explore the pruning algorithm to increase the sparsity levels. The research will also concentrate on adapting the pruning as a neuroevolution search algorithm.

\section*{Symbols}

\begin{itemize}
    \item $N$ -- number of samples in the batch - batch size,
    \item $top1$ -- top1 accuracy,
    \item $mAP$ -- mean average precision,
    \item $F1$ -- F1 metric
    \item $TP$ -- number of correctly detected anomalies
    \item $FN$ -- anomalies mistakenly classified as normal data point
    \item $FP$ -- data points mistakenly classified as anomalies
    \item $X$ -- input data to the model
    \item $A$ -- activation values
    \item $F_{\Theta}^p$ -- pruned model
    \item $F_{\Theta}$ -- original dense model
    \item $f_{\theta_i}$ -- convolutional or fully-connected layer
    \item $M$ -- mask of the model
    \item $G$ -- model's gradient statistics
    \item $O$ -- model optimizer
    \item $\Theta$ -- set of weights in model layers 
    \item $g_i$ -- sum of weight and average gradient absolute magnitude
    \item $C_i, D_i$ -- numbers of input and output channels of the layer $i$, 
    \item $H_i, W_i$ -- height and width of the filter of layer $i$,
    \item $\omega$ -- number of cluster in nonlinear quantisation
    \item $\psi$ -- centroid bit-width quantisation
    \item $C$ -- list of centroids
    \item $\theta_{ca}$ -- list of weight to centroid assignments
    \item $\Delta_i$ -- sparsity update factor for the layer $i$
    \item $\alpha$ -- coefficient for indicator $g_{\theta_i}$ computation
    \item $\alpha_m$ -- coefficient for scaling the update of sparsity in mutation operator
    \item $\gamma$ -- probability threshold for genetic operations
    \item $\epsilon$ -- accuracy threshold in pruning algorithm
    \item $\epsilon_S$ -- threshold for $g_{\theta_i}$ magnitude
    \item $\beta$ -- scaling factor for $\Delta_i$ parameter
    \item $w$ -- single weight
    \item $w_q$ -- single quantized weight
    \item $R$ -- ranked list
    \item $R_L$ -- ranked list length
    \item $N_{it}$ -- number of pruning algorithm iterations
    \item $S$ -- number of weights in the model
    \item $P$ -- number of output neurons in a layer
    \item $S_{\theta_{i}}$ -- number of weights in the layer $i$
    \item $SH_{\theta_{i}}$ -- shape of weights tensor in the layer $i$
    \item $B$ -- number of batches
    \item $L$ -- number of convolutional and fully-connected layers in the model
    \item $cross$ -- crossover function
    \item $\sigma, \mu$ -- linear quantisation parameters
    \item $t_{in_{p}}$ -- address of input data which is copied to thread local memory 
    \item $N_B$ -- number of thread blocks on GPU
    \item $X_w$, $Y_w$ -- width of input and output data 
    \item $X_{size}$ -- size of one sample of the input data
    \item $s_h, s_w$ -- height and width of stride
    \item $h_{pad}, w_{pad}$ -- height and width of padding
    \item $t_{out_x}, t_{out_y}$, $t_{out_z}$ -- CUDA thread x, y and z coordinates 
    \item $sb_S$ -- number of input vector which are loaded into single threads block
    \item $N_{nz}$ -- number of non zero elements in each output channel
    \item $RP_i$ -- row pointer structure for layer $i$
    \item $\Lambda_i$ -- array of CSR format, stores indexes of nonzero elements in layer $i$
    \item $sh_{idx}$ -- index of $\theta_i$ and $\Lambda_i$ loaded to thread block
\end{itemize}

\bibliographystyle{unsrt}  


\bibliography{references}
\end{document}